\pdfoutput=1
\documentclass[11pt,a4paper]{article}
\PassOptionsToPackage{linktocpage}{hyperref}
\usepackage[hyperref]{emnlp2018}
\usepackage{times}
\usepackage{latexsym}
\usepackage{todonotes}
\usepackage{amsmath,amssymb}

\usepackage{url}

\usepackage{booktabs}

\usepackage{enumitem}

\newcommand{\relu}{\textsf{relu}}
\newcommand{\lrelua}{\textsf{lrelu-0.01}}
\newcommand{\lrelub}{\textsf{lrelu-0.30}}
\newcommand{\swish}{\textsf{swish}}
\newcommand{\pentan}{\textsf{penalized tanh}}
\newcommand{\arctid}{\textsf{arctid}}
\newcommand{\elu}{\textsf{elu}}
\newcommand{\minsin}{\textsf{minsin}}
\newcommand{\maxtanh}{\textsf{maxtanh}}
\newcommand{\mytanh}{\textsf{tanh}}
\newcommand{\mysin}{\textsf{sin}}
\newcommand{\maxsig}{\textsf{maxsig}}
\newcommand{\selu}{\textsf{selu}}
\newcommand{\linear}{\textsf{linear}}
\newcommand{\sigmoid}{\textsf{sigmoid}}
\newcommand{\cosid}{\textsf{cosid}}
\newcommand{\cube}{\textsf{cube}}
\newcommand{\maxouta}{\textsf{maxout-2}}
\newcommand{\maxoutb}{\textsf{maxout-3}}
\newcommand{\maxoutc}{\textsf{maxout-4}}
\newcommand{\prelu}{\textsf{prelu}}
\newcommand{\best}{\texttt{best}}
\newcommand{\avg}{\texttt{mean}}

\aclfinalcopy 
\def\aclpaperid{256} 

\title{Is it Time to Swish? \\ Comparing Deep Learning Activation Functions Across NLP tasks}

\author{Steffen Eger, Paul Youssef, Iryna Gurevych \\
  Ubiquitous Knowledge Processing Lab (UKP-TUDA) \\
  Department of Computer Science \\
  Technische Universit\"at Darmstadt\\
  {\tt www.ukp.tu-darmstadt.de} 
  }

\date{}

\begin{document}
\maketitle
\begin{abstract}
Activation functions play a crucial role in neural networks because they are the non-linearities which have been attributed to the success story of deep learning. One of the currently most popular activation functions is ReLU, but several competitors have recently been proposed or `discovered', including LReLU functions and \swish. While most works compare newly proposed activation functions on few tasks (usually from image classification) and against few competitors (usually ReLU), we perform the first large-scale comparison of 21 activation functions across eight different NLP tasks. We find that a largely unknown activation function performs most stably across all tasks, the so-called \pentan{} function. We also show that it can successfully replace the \sigmoid{} and \mytanh{} gates in LSTM cells, leading to a 2 percentage point (pp) improvement over the standard choices on a challenging NLP task. 
\end{abstract}

\section{Introduction}\label{sec:introduction}
Activation functions are a crucial component of neural networks
because they turn an otherwise linear classifier into a non-linear one, which has proven key to the high performances witnessed across a wide range of tasks in recent years. While different activation functions such as \sigmoid{} or \mytanh{} are often equivalent on a theoretical level, in the sense that they can all approximate arbitrary continuous functions \cite{Hornik:1991}, different activation functions often show very diverse behavior in practice.

For example, \sigmoid{}, one of the activation functions dominating in neural network practice for several decades
eventually turned out less suitable for learning because (according to accepted wisdom) of its small derivative which may lead to vanishing gradients. 
In this respect, the so-called ReLU function \cite{Glorot:2011} has proven much more suitable.
It has an identity derivative in the positive region and is thus claimed to be less susceptible to vanishing gradients. It has therefore (arguably) become the most popular activation function. The recognition of ReLU's success has led to various extensions proposed \cite{Maas:2013,He:2015,Klambauer:2017}, but none has reached the same popularity, most likely because of ReLU's simplicity and because the gains reported tended to be inconsistent or marginal across datasets and models \cite{Ramach:2018}.

Activation functions have been characterized by a variety of properties deemed important for successful learning, such as ones 
relating to their derivatives, monotonicity, and whether their range is finite or not. However, in recent work, \citet{Ramach:2018} employed automatic search to find high-performing novel activation functions, where their search space contained compositions of elementary unary and binary functions such as $\max$, $\min$, $\sin$, $\tanh$, or $\exp$. They found many functions violating properties deemed as useful, such as non-monotonic activation functions or functions violating the gradient-preserving property of ReLU. Indeed, their most successful function, which they call \swish{}, violates both of these conditions. However, as with previous works, they also only evaluated their newly discovered as well as their (rectifier) baseline activation functions on few different datasets, usually taken from the image classification community such as CIFAR \cite{Krizhevsky:2009} and ImageNet \cite{Russakovsky:2015}, and using few types of different networks, such as the deep convolutional networks abounding in the image classification community \cite{Szegedy:2016}.

To our best knowledge, there exists no large-scale empirical comparison of different activations across a variety of tasks and network architectures, and even less so within natural language processing (NLP).\footnote{An exception may be considered \citet{Xu:2015}, who, however, only contrast the rectifier functions on image classification datasets.} Thus, the question which activation function really performs best and most stably across different NLP tasks and popular NLP models remains unanswered to this date. 

In this work, we fill this gap. We compare (i) 21 different activation functions, including the 6 top performers found from automatic search in \citet{Ramach:2018}, across (ii) three popular NLP task types (sentence classification, document classification, sequence tagging) comprising 8 individual tasks, (iii) using three different popular NLP architectures, namely, MLPs, CNNs, and RNNs. We also (iv) compare all functions across two different dimensions, namely: top vs.\ average performance. 

We find that 
a largely unknown activation function, \pentan{} \cite{Xu:2016}, performs most stably across our different tasks. We also find that it can successfully replace \mytanh{} and \sigmoid{} activations in LSTM cells. We further find that the majority of top performing functions found in \citet{Ramach:2018} do not perform well for our tasks. An exception is \swish{}, which performed well across several tasks, but less stably than \pentan{} and other functions.\footnote{Accompanying code to reproduce our experiments is available from \url{https://github.com/UKPLab/emnlp2018-activation-functions}.}


\section{Theory}\label{sec:theory}
\paragraph{Activation functions} We consider 21 activation functions, 6 of which are ``novel'' and proposed in \citet{Ramach:2018}. The functional form of these 6 is given in Table \ref{table:functions}, together with the \sigmoid{} function. 
\begin{table}[!htb]
  \centering
  \begin{tabular}{ll}
  \toprule
    \sigmoid & $f(x)=\sigma(x)=1/(1+\exp(-x))$\\
    \swish & $f(x)=x\cdot \sigma(x)$\\
    \maxsig & $f(x)=\max\{x,\sigma(x)\}$\\
    \cosid & $f(x)=\cos(x)-x$\\
    \minsin & $f(x)=\min\{x,\sin(x)\}$\\
    \arctid & $f(x)=\arctan(x)^2-x$\\
    \maxtanh & $f(x)=\max\{x,\tanh(x)\}$\\
    \midrule
    \lrelua & $f(x)=\max\{x,0.01x\}$ \\
    \lrelub & $f(x)=\max\{x,0.3x\}$ \\
    {\small \pentan} & $f(x)=\begin{cases}\tanh(x) & x>0,\\ 0.25\tanh(x) & x\le 0\end{cases}$\\
    \bottomrule
  \end{tabular}
  \caption{Top: \sigmoid{} activation function as well as 6 top performing activation functions from \citet{Ramach:2018}. Bottom: the LReLU functions with different parametrizations as well as \pentan{}.}
  \label{table:functions}
\end{table}

The remaining 14 are: \mytanh, \mysin, \relu, \lrelua, \lrelub, \maxouta, \maxoutb, \maxoutc, \prelu, \linear, \elu{}, \cube, \pentan, \selu{}. We briefly describe 
them: 
\lrelua{} and \lrelub{} are the so-called leaky relu (LReLU) functions \cite{Maas:2013}; the idea behind them is to avoid zero activations/derivatives in the negative region of \relu{}. Their functional form is given in Table \ref{table:functions}. \prelu{} \cite{He:2015} generalizes the LReLU functions by allowing the slope in the negative region to be a 
learnable parameter. The maxout functions \cite{Goodfellow:2013} are different in that they introduce additional parameters and do not operate on a single scalar input. For example, \maxouta{} is the operation that takes the maximum of two inputs: $\max\{\mathbf{xW}+\mathbf{b},\mathbf{xV}+\mathbf{c}\}$, so the number of learnable parameters is doubled. \maxoutb{} is the analogous function that takes the maximum of three inputs.
As shown in \citet{Goodfellow:2013}, maxout can approximate any convex function. \mysin{} is the standard sine function, proposed in neural network learning, e.g., in \citet{Parascandolo:2016}, where it was shown to enable faster learning on certain tasks than more established functions. \pentan{} \cite{Xu:2016} has been defined in analogy to the LReLU functions, which can be thought of as ``penalizing'' the identity function in the negative region. 
The reported good performance of \pentan{} on CIFAR-100 \cite{Krizhevsky:2009}
lets the authors speculate that the slope 
of activation functions near the origin may be crucial for learning. 
\linear{} is the identity function, $f(x)=x$. \cube{} is the function $f(x)=x^3$, proposed in \citet{Chen:2014} for an MLP used in dependency parsing. \elu{} \cite{Clevert:2015} has been proposed as (yet another) variant of \relu{} that assumes negative values, making the mean activations more zero-centered. \selu{} is a scaled variant of \elu{} used in \citet{Klambauer:2017} in the context of so-called self-normalizing neural nets.

\paragraph{Properties of activation functions} 
\begin{table*}[!htb]
\centering
\footnotesize
\begin{tabular}{llll}
  \toprule
  Property & Description & Problems & Examples \\ \midrule
  derivative & $f'$ & $>1$ exploding gradient (e) &  \sigmoid{} (v), \mytanh{} (v), \cube{} (e)\\
  & & $<1$ vanishing (v) & \\
  zero-centered & range centered around zero? &   if not, slower learning & \mytanh{} ($+$), \relu{} ($-$) \\ 
  saturating & finite limits & vanishing gradient in the limit & \mytanh{}, \pentan{}, \sigmoid{}\\
  monotonicity & $x>y\implies f(x)\ge f(y)$ & unclear & exceptions: \mysin{}, \swish{}, \minsin{} 
  \\ \bottomrule
 \end{tabular}
 \caption{Frequently cited properties of activation functions}.
 \label{table:properties}
\end{table*}

Many properties of activation functions have been speculated to be crucial for successful learning. 
Some of these are listed in Table \ref{table:properties}, together with brief descriptions and illustrations. 

Graphs of all activation functions can be found in the appendix. 

\section{Experiments}\label{sec:experiments}
We conduct experiments using three neural network types and three types of NLP tasks, described in \S\ref{sec:1}, \S\ref{sec:2}, and \S\ref{sec:3} below.
\subsection{MLP \& Sentence Classification}\label{sec:1}
\paragraph{Model}
We experiment with a multi-layer perceptron (MLP) applied to sentence-level classification tasks. That is, input to the MLP is a sentence or short text, represented as a fixed-size vector embedding. The output of the MLP is a label which classifies the sentence or short text. We use two sentence representation techniques, namely, Sent2Vec \cite{Pagliardini:2018}, of dimensionality 600, and InferSent \cite{Conneau:2017}, of dimensionality 4096. Our MLP has the form:
\begin{align*}
  \mathbf{x}_i &= f(\mathbf{x}_{i-1}\cdot \mathbf{W}_i+\mathbf{b}_i)\\
  \mathbf{y} &= \text{softmax}(\mathbf{x}_{N}\mathbf{W}_{N+1}+\mathbf{b}_{N+1})
\end{align*}
where $\mathbf{x}_0$ is the input representation, $\mathbf{x}_1,\ldots,\mathbf{x}_{N}$ are hidden layer representations, and $\mathbf{y}$ is the output, a probability distribution over the classes in the classification task. Vectors $\mathbf{b}$ and matrices $\mathbf{W}$ are the learnable parameters of our network. The activation function is given by $f$ and ranges over the choices described in \S\ref{sec:theory}. 
\paragraph{Data}
We use four sentence classification tasks, namely: movie review classification (MR), subjectivitiy classification (SUBJ), 
question type classification (TREC), 
and classifying whether a sentence contains an argumentation structure of a certain type (claim, premise, major claim) or else is non-argumentative (AM). The first three datasets are standard sentence classification datasets and contained in the SentEval framework.\footnote{\url{https://github.com/facebookresearch/SentEval}} We choose the AM dataset for task diversity, and derive 
it by projecting token-level annotations in the dataset from \citet{Stab:2017} to the sentence level. In the rare case ($<$5\% of the cases) when a sentence contains multiple argument types, we choose one based on the ordering Major Claim (MC) $>$ Claim (C) $>$ Premise (P). Datasets and examples are listed in Table \ref{table:data_sent}.

\begin{table*}[!htb]
\centering
\small
\begin{tabular}{llccll}
\toprule
\textbf{Task} & \textbf{Type} & \textbf{Size} & \textbf{C} & \textbf{Example} \\ 
\midrule
{AM} & Argumentation & 7k  & 4 &
Not cooking fresh food will lead to lack of nutrition. \textit{(claim)} 
 \\
{MR} & Sentiment & 11k & 2 & Too slow for a younger crowd , too shallow for an older one. \textit{(neg)} \\
{SUBJ} & Subjectivity & 10k & 2 & A movie that doesn’t aim too high , but doesn’t need to. \textit{(subj)} \\
{TREC} & Question-types & 6k & 6 & What's the Olympic Motto? \textit{(description)} \\
\midrule
NG & Doc classification & 18k & 20 & [...] You can add "dark matter" and quarks [...] (\emph{sci.space})\\
R8 & Doc classification & 7k & 8 & bowater industries profit exceed [...] (\emph{earn})\\
\midrule
POS & POS tagging & 204k & 17 & What/\emph{PRON} to/\emph{PART} feed/\emph{VERB} my/\emph{PRON} dog/\emph{NOUN} [...]\\
TL-AM & Token-level AM & 148k & 7 & [...] I/\emph{O} firmly/\emph{O} believe/\emph{O} that/\emph{O} we/\emph{B-MC} should/\emph{I-MC} [...]\\
\bottomrule
\end{tabular}
\caption{Evaluation tasks used in our experiments, grouped by task type (sentence classification, document classification, sequence tagging), with statistics and examples. C is the number of classes to predict.}
\label{table:data_sent}
\end{table*}

\paragraph{Approach}
We consider 7 ``mini-experiments'':
\begin{itemize}[noitemsep,leftmargin=0.6cm]
  \item (1): MR dataset with Sent2Vec-unigram embeddings as input and 1\% of the full data as training data; (2): the same mini-experiment with 50\% of the full data as training data. In both cases, the dev set comprises 10\% of the full data and the rest is for testing. 
  \item (3,4): SUBJ with InferSent embeddings and likewise 1\% and 50\% of the full data as train data, respectively.
  \item (5): The TREC dataset with original split in train and test; 50\% of the train split is used as dev data. 
  \item (6): The AM dataset with original split in train, dev, and test \cite{Eger:2017}, and with InferSent input embeddings. (7): the same mini-experiment with Sent2Vec-unigram embeddings. 
\end{itemize}
We report accuracy for mini-experiments (1-5) and macro-F1 for (6-7). We report macro-F1 for (6-7) because the AM dataset is imbalanced.  

The motivation behind choosing different input embeddings for different tasks was to investigate a wider variety of conditions. 
Choosing subsets of the full data had the same intention. 

For all 7 mini-experiments, we draw the same 200 randomly chosen hyperparameters from the ranges indicated in Table \ref{table:hyperparams_sent}. 
All experiments are conducted in keras.\footnote{\url{https://keras.io/}} 

For each of the 21 different activation functions detailed in \S\ref{sec:theory}, we conduct each mini-experiment with the 200 randomly chosen hyperparameters. All activation functions use the same hyperparameters and the same train, dev, and test splits.


We store two results for each mini-experiment, namely: (i) the test result corresponding to the \textbf{best} (\best) dev performance; (ii) the \textbf{average} (\avg) test result across all hyperparameters. The \best{} result scenario mirrors standard optimization in machine learning: it indicates the score one can obtain with an activation function when 
the MLP is well-optimized. The \avg{} result scenario is an indicator for what one can expect when hyperparameter optimization is `shallow' (e.g., because computing times are prohibitive): it gives the average performance for randomly chosen hyperparameters. We note that we run each hyperparameter combination 
with 5 different random weight initializations and all the reported scores (best dev score, best \best{}, best \avg{}) are averages over these 5 random initializations. 

Finally, we set the following hyperparameters for all MLP experiments: patience of 10 for early stopping, batch size 16, 100 epochs for training. 

\begin{table*}[htb]
  \centering
  {\small
  \begin{tabular}{llr}
  \toprule
    Model & Hyperparameter & Range \\ \toprule
    (a) MLP & (1) optimizer & $\{$Adam,RMSprop,Adagrad,Adadelta,Adamax,Nadam,sgd$\}$ \\
    & (2) \#hidden layers $N$ & $\{1,2,3,4\}$\\
    & (3) dropout value & $[0.1,0.75]$ \\
    & (4) hidden units & $[30,500]$ \\
    & (5) learning rate & $\mathcal{N}(m,m/5)$\\
    & (6) weight initializer & $\{$random-n, random-u, varscaling, orthogonal, \\
    & & lecun-u, glorot-n, glorot-u, he-n, he-u$\}$\\
    \midrule
     (b) CNN        & (a) (1,3,5,6) & same as MLP \\
             & embedding dimension & $[40,200]$\\
     & number of filters $n_k$ & $[30,500]$\\
             & \#hidden layers $N$ & $\{1,2,3\}$\\
             & filter size $h$ & $\{1,2,2,3,3,3,4\}$\\
    \midrule
      (c) RNN/LSTM & (a) (1-5) & same as MLP \\
     & recurrent initializer & same as (a) (6) plus identity matrix\\
    \midrule
  \end{tabular}
  \caption{Hyperparameter ranges for each network type. Hyperparameters are drawn using a discrete or continuous uniform distribution from the indicated ranges. Repeated values indicate multi-sets. $\mathcal{N}(\mu,s)$ is the normal distribution with mean $\mu$ and std $s$; $\mu=m$ is the default value from keras for the specific optimizer (if drawn learning rate is $<0$, we choose it to be $m$).}
  \label{table:hyperparams_sent}
  }
\end{table*}

\begin{figure}[htb]
  \centering
  \scalebox{0.5}{\input{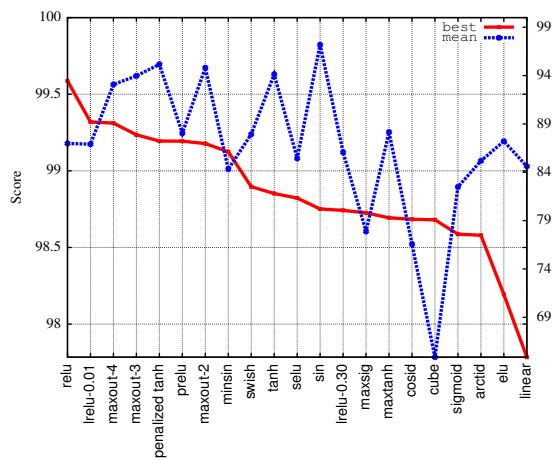}}
  \caption{Sentence Classification. Left y-axis: \best. Right y-axis: \avg{}. Score on y-axes is the average over all mini-experiments.}
  \label{fig:sent}
\end{figure}

\paragraph{Results}
Figure \ref{fig:sent} shows \best{} and \avg{} results, averaged over all 7 mini-experiments, for each activation function. To make individual scores comparable across mini-experiments, we perform max normalization and divide each score by the maximum score achieved in any given mini-experiment (for \best{} and \avg{}, respectively) before averaging.\footnote{We chose max normalization so that certain tasks/mini-experiments would not unduly dominate our averaged scores. Overall, our averaged scores are not (much) affected by this decision, however: the Spearman correlation of rankings of activation functions under max normalization and under no max normalization are above 0.98 in all our three classification scenarios considered in \S\ref{sec:1},\S\ref{sec:2},\S\ref{sec:3}.}

For \best{}, the top performers are the rectifier functions (\relu{}, \lrelua{}, \prelu{}) as well as maxout and \pentan{}. The newly discovered activation functions lag behind, with the best of them being \minsin{} and \swish{}. \linear{} is worst, together with \elu{} and \cube. Overall, 
the difference between the best activation function, \relu{}, and the worst, \linear, is only roughly 2pp, however. This means that if hyperparameter search is done carefully,  the choice of activation function is less important for these sentence classification tasks. Particularly the (binary) tasks MR and SUBJ appear robust against the choice of activation function, with the difference between the best and worst function being always less than 1pp, in all settings. For TREC and AM, the situation is slightly different: for TREC, the difference is 2pp (\swish{} vs.\ \maxsig) and for AM, it is 3pp using InferSent embeddings (\swish{} vs.\ \cube) and 12pp using Sent2Vec embeddings (\relu{} vs.\  \linear). It is noteworthy that \swish{} wins 2 out of 3 cases in which the choice of activation function really matters. 

\avg{} results are very different from \best{} results. Here, somewhat surprisingly, the oscillating \mysin{} function wins, followed by \pentan{}, maxout and \swish. The difference between the best \avg{} function, \mysin, and the worst, \cube, is more than 30pp. This means that using \cube{} is much riskier and requires more careful hyperparameter search compared to \mysin{} and the other top performers. 

\subsection{CNN \& Document Classification}\label{sec:2}
\paragraph{Model} Our second paradigm is document classification using a CNN. This approach has been popularized in NLP by the ground-breaking work of \citet{Kim:2014}. Even though shallow CNNs do not reach state-of-the-art results on large datasets anymore \cite{Johnson:2017}, simple approaches like (shallow) CNNs are still very competitive for smaller datasets \cite{Joulin:2016}. 

Our model operates on token-level and first embeds a sequence of tokens $x_1,\ldots,x_n$, represented as 1-hot vectors, into learnable embeddings $\mathbf{x}_1,\ldots,\mathbf{x}_n$. The model then applies 1D-convolution on top of these embeddings. That is, a filter $\mathbf{w}$ of size $h$ takes $h$ successive embeddings $\mathbf{x}_{i:i+h-1}$, performs a scalar product and obtains a feature $c_i$:
\begin{align*}
  c_i = f(\mathbf{w}\cdot \mathbf{x}_{i:i+h-1}+b).
\end{align*}
Here, $f$ is the activation function and $b$ is a bias term. We take the number $n_k$ of different filters as a hyperparameter. When our network has multiple layers, we stack another convolutional layer on top of the first (in total we have $n_k$ outputs at each time step), and so on. Our penultimate layer is a global max pooling layers that selects the maximum from each feature map. A final softmax layer terminates the network. 
\paragraph{Data} We use two document classification tasks, namely: 20 Newsgroup (NG) and Reuters-21578 R8 (R8). Both datasets are standard document classification datasets. In NG, the goal is to classify each document into one of 20 newsgroup classes (alt.atheism, sci.med, sci.space, etc.). In R8, the goal is to classify Reuters news text into one of eight classes (crude, earn, grain, interest, etc.).  We used the preprocessed files from \url{https://www.cs.umb.edu/~smimarog/textmining/datasets/} (in particular, stopwords are removed and the text is stemmed).
\paragraph{Approach}
We consider 4 mini-experiments:
\begin{itemize}[noitemsep,leftmargin=0.6cm]
\item (1,2) NG dataset with 5\% and 50\%, respectively of the full data as train data. In both cases, 10\% of the full data is used as dev data, and the rest as test data.  
\item (3,4) Same as (1,2) for R8. 
\end{itemize}
We report accuracy for all experiments. We use a batch size of 64, 50 epochs for training, and a patience of 10. For all mini-experiments, we again draw 200 randomly chosen hyperparameters from the ranges indicated in Table \ref{table:hyperparams_sent}. 
The hyperparameters and train/dev/test splits are the same for all activation functions. 

\begin{figure}[!htb]
\centering
\scalebox{0.5}{\input{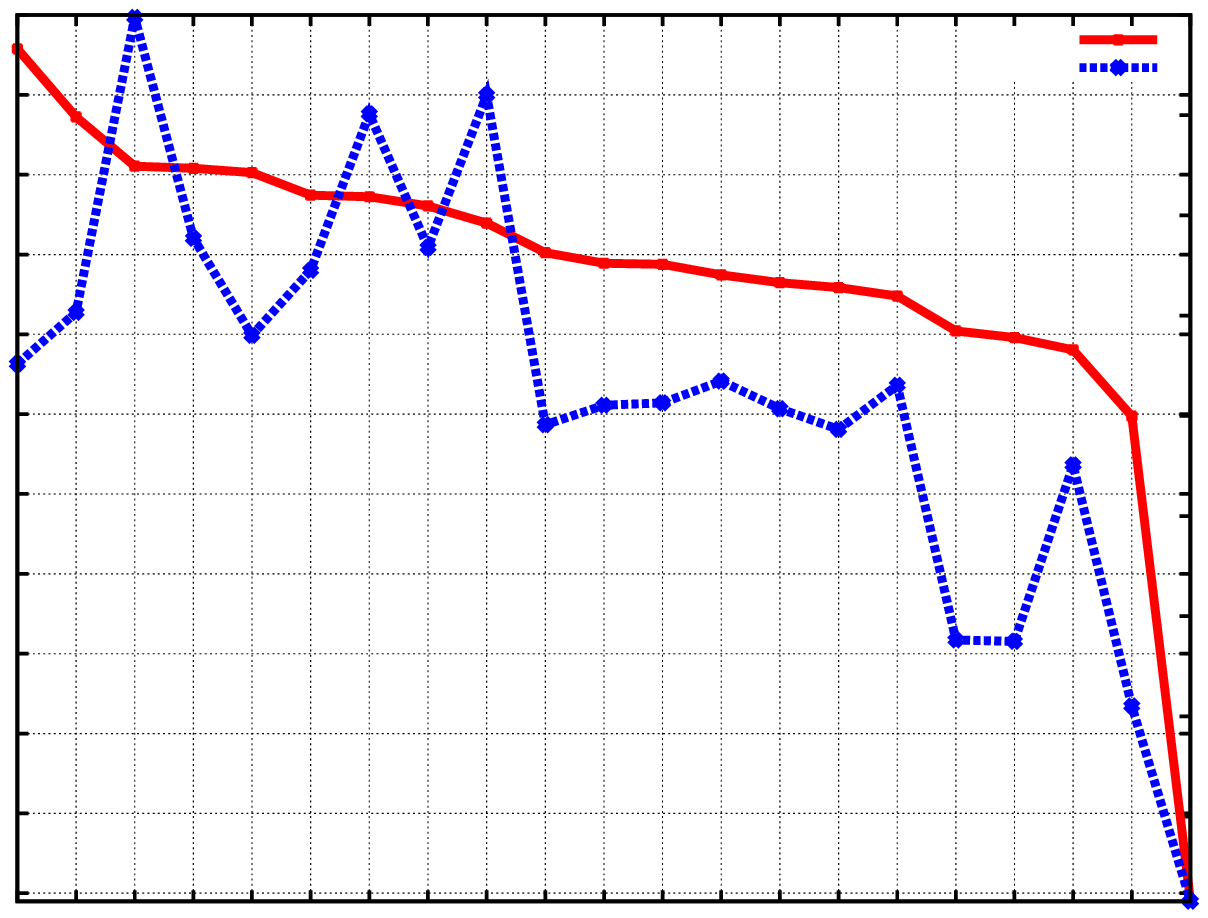}}
\caption{Doc classification.}
\label{fig:doc}
\end{figure}

\paragraph{Results}
Figure \ref{fig:doc} shows \best{} and \avg{} results, averaged over all mini-experiments. This time, the winners for \best{} are \elu{}, \selu{} (again two members from the rectifier family), and \maxoutb{}, but the difference between \maxoutb{} and several lower ranked functions is minimal. The \cube{} function is again worst and \sigmoid{} and \cosid{} have similarly bad performance. Except for \minsin{}, the newly proposed activation functions from \citet{Ramach:2018} again considerably lag behind. The most stable activation functions are the maxout functions as well as \pentan{}, \mytanh{} and \mysin{}.  

\subsection{RNN \& Sequence Tagging}\label{sec:3}
\paragraph{Model} Our third paradigm is sequence tagging, a ubiquitous model type in NLP. In sequence tagging, a sequence of input tokens $w_1,\ldots,w_K$ is mapped to a sequence of labels $y_1,\ldots,y_K$. Classical sequence tagging tasks include POS tagging, chunking, NER, discourse parsing \cite{Braud:2017}, and argumentation mining \cite{Eger:2017,Schulz:2018}. We use a standard recurrent net for sequence tagging, whose form is:
\begin{align*}
  \mathbf{h}_i &= f(\mathbf{h}_{i-1}\mathbf{W}+\mathbf{w}_i\cdot\mathbf{U}+\mathbf{b})\\
  \mathbf{y}_i &= \text{softmax}(\mathbf{h}_i\mathbf{V}+\mathbf{c})
\end{align*}
Here, $\mathbf{w}_i$ are (pre-trained) word embeddings of words $w_i$. Vectors $\mathbf{b},\mathbf{c}$ and matrices $\mathbf{U},\mathbf{V},\mathbf{W}$ are parameters to be learned during training. The above describes an RNN with only one hidden layer, $\mathbf{h}_i$, at each time step, but we consider the generalized form with $N\ge 1$ hidden layers; we also choose a bidirectional RNN in which the hidden outputs of a forward RNN and a backward RNN are combined. 
RNNs are particularly deep networks---indeed, the depth of the network corresponds to the length of the input sequence---which makes them particularly susceptible to the vanishing gradient problem \cite{Pascanu:2013}. 

Initially, we do not consider the more popular LSTMs here for reasons indicated below. However, we include a comparison after discussing the RNN performance. 

\paragraph{Data}
We use two sequence tagging tasks, namely: English POS tagging (POS), and token-level argumentation mining (TL-AM) using the same dataset (consisting of student essays) as for the sentence level experiments. In token-level AM, we tag each token with a BIO-label plus the component type, i.e., the label space is $\mathcal{Y}=\{\text{B},\text{I}\}\times\{\text{MC},\text{C},\text{P}\}\cup\{\text{O}\}$, where 
`O'' is a label for non-argumentative tokens. The motivation for using TL-AM is that, putatively, AM has more long-range dependencies than POS or similar sequence tagging tasks such as NER, because argument components are much longer than named entities and component labels also 
depend less 
on the current token.  

\begin{figure}[!htb]
\centering
\scalebox{0.5}{
\input{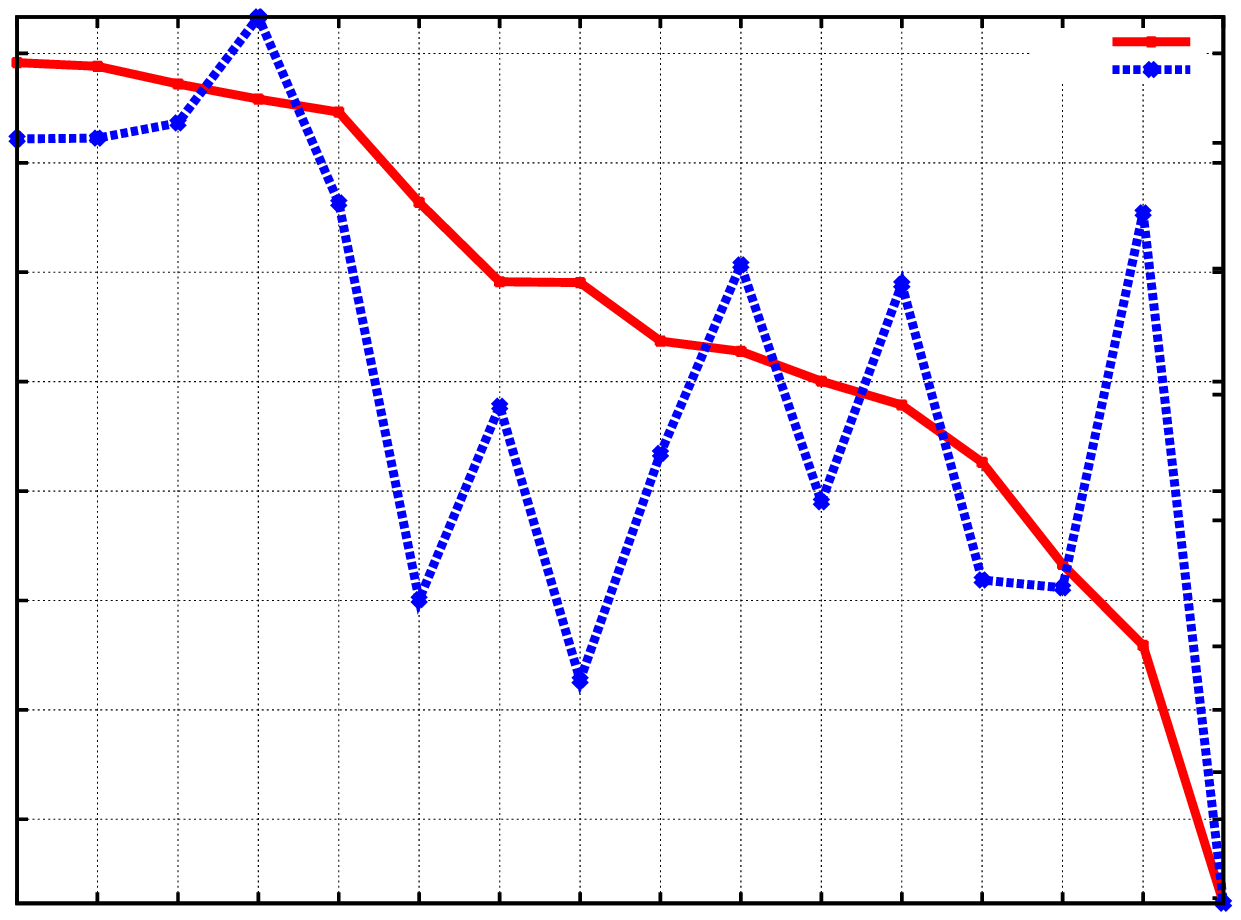}}
\caption{Sequence tagging.}
\label{fig:seq}
\end{figure}

\paragraph{Approach} We consider 6 mini-experiments: 
\begin{itemize}[noitemsep,leftmargin=0.6cm]
  \item (1): TL-AM with Glove-100d word embeddings and 5\% of the original training data as train data; (2) the same with 30\% of the original training data as train data. In both cases, dev and test follow the original train splits \cite{Eger:2017}. 
  \item (3,4) Same as (1) and (2) but with 300d Levy word embeddings \cite{Levy:2014}.
  \item (5,6): POS with Glove-100d word embeddings and 5\% and 30\%, respectively, of the train data of a pre-determined train/dev/test split (13k/13k/178k tokens). Dev and test are fixed in both cases.    
\end{itemize}
We report macro-F1 for mini-experiments (1-4) and accuracy for (5-6). For our RNN implementations, we use the accompanying code of (the state-of-the-art model of) \citet{Reimers:2017}, which is implemented in keras. The network uses a CRF layer as an output layer. We use a batch size of 32, train for 50 epochs and use a patience of 5 for early stopping.
\paragraph{Results}
Figure \ref{fig:seq} shows \best{} and \avg{} results, averaged over all 6 mini-experiments, for each activation function. We exclude \prelu{} and the maxout functions because the keras implementation does not natively support these activation functions for RNNs. We also exclude the \cube{} function because it performed very badly.  

Unlike for sentence classification, there are much larger differences between the activation functions. For example, there is almost 20pp difference between the best \best{} activation functions: \relu{}, \lrelua{}, \swish{}, \pentan{}, and the worst ones: \linear{}, \cosid{}, and \sigmoid{} (the differences were larger had we included \cube{}). Interestingly, this difference is mostly due to the TL-AM task: for POS, there is only 3pp difference between the best function (\sigmoid{} (sic!), though with almost zero margin to the next best ones) and the worst one (\linear), while this difference is almost 40pp for TL-AM. 
This appears to confirm our concerns regarding the POS tagging task as not being challenging enough due to lack of, e.g., long-range dependencies.

The four best \best{} activation functions in Figure \ref{fig:seq} are also the functions with the best \avg{} results, i.e., they are most stable over different hyperparameters. 
The clear winner in this category is \pentan{} with 100\% \avg{} score, followed by \swish{} with 91\%. Worst is \cosid{} with 30\%. It is remarkable how large the difference between \mytanh{} and \pentan{} is both for \best{} and \avg{}---7pp and 20pp, respectively, which is much larger than the differences between the analogous pair of LReLU and \relu{}. This appears to make a strong case for the importance of the slope around the origin, as suggested in \citet{Xu:2016}. 


\paragraph{LSTM vs.\ RNN} Besides an RNN, we also implemented a more popular RNN model with (bidirectional) LSTM blocks in place of standard hidden layers. Standard LSTM units follow the equations (simplified):
\begin{align*}
  \mathbf{f}_t &= \sigma([\mathbf{h}_{t-1};\mathbf{x}_t]\cdot \mathbf{W}_f),\\
  \mathbf{i}_t &= \sigma([\mathbf{h}_{t-1};\mathbf{x}_t]\cdot \mathbf{W}_i),\\
  \mathbf{o}_t &= \sigma([\mathbf{h}_{t-1};\mathbf{x}_t]\cdot \mathbf{W}_o)\\
  \mathbf{c}_t &= \mathbf{f}_t\odot \mathbf{c}_{t-1}+\mathbf{i}_t\odot\tau([\mathbf{h}_{t-1};\mathbf{x}_t]\cdot \mathbf{W}_c)\\
  \mathbf{h}_t &= \mathbf{o}_t\odot \tau(\mathbf{c}_t),
\end{align*}
where $\mathbf{f}_t$ and $\mathbf{i}_t$ are perceived of as \emph{gates} that control information flow,
$\mathbf{x}_t$ is the input at time $t$ and $\mathbf{h}_t$ is the hidden layer activation. In keras (and most standard references), $\sigma$ is the (hard) sigmoid function, and $\tau$ is the \mytanh{} function. 

We ran an LSTM on the TL-AM dataset with Levy word embeddings and 5\% and 30\% data size setup. We varied $\sigma$ and $\tau$ independently, keeping the respective other function at its default. 

We find that the top two choices for $\tau$ are \pentan{} and \mytanh{} (margin of 10pp), given that $\sigma$ is sigmoid. For $\tau=$ \mytanh{}, the best choices are $\sigma=$ \pentan{}, \sigmoid, and \mytanh{}. All other functions perform considerably worse. Thus, the top-performers are all saturating functions, indicating the different roles activation functions play in LSTMs---those of gates---compared to standard layers. It is worth mentioning that choosing $\sigma$ or $\tau$ as \pentan{} is on average better than the standard choices for $\sigma$ and $\tau$. 
Indeed, choosing $\tau=\sigma=$ \pentan{} is on average 2pp better than the default choices of $\tau,\sigma$.

It is further worth mentioning that the best \best{} results for the LSTM are roughly 5pp better (absolute) than the best corresponding choices for the simple RNN.

\section{Analysis \& Discussion}\label{sec:analysis}
\paragraph{Winner statistics} 
Each of the three meta-tasks sentence classification, document classification, and sequence tagging was won, on average, by a member from the rectifier family, namely, \relu{} (2) and \elu{}, for \best{}. Also, in each case, \cube{} and \cosid{} were among the worst performing activation functions. The majority of newly proposed functions from \citet{Ramach:2018} ranked somewhere in the mid-field, with \swish{} and \minsin{} performing best in the \best{} category.
For the \avg{} category, we particulary had the maxout functions as well as \pentan{} and \mysin{} regularly as top performers. 

To get further insights, we computed a winner statistic across all 17 mini-experiments, counting how often each activation function was among the top 3. Table \ref{table:topN} shows the results, excluding \prelu{} and the maxout functions because they were not considered in all mini-experiments. 

\begin{table}[!htb]
  \centering
  \begin{tabular}{ll}
  \toprule
    \best{} & \pentan{} (6), \swish{} (6), \\
    & \elu{} (4), \relu{} (4), \lrelua{} (4)\\   
    \avg{} & \pentan{} (16), \mytanh{} (13)\\
    & \mysin{} (10) \\
  \bottomrule
  \end{tabular}
  \caption{Top-3 winner statistics. In brackets: number of times within top-3, keeping only functions with four or more top-3 rankings.}
  \label{table:topN}
\end{table}

We see that \pentan{} and \swish{} win here for \best, followed by further rectifier functions. The \avg{} category is clearly won by saturating activation functions with finite range. If this comparison were restricted to sentence and document classification, where we also included the maxout functions, then \pentan{} would have been outperformed by maxout for \avg{}.

This appears to yield the conclusion that functions with limited range behave more stably across hyperparameter settings while non-saturating functions tend to yield better top-performances. The noteworthy exception to this rule is \pentan{} which excels in both categories (the more expensive maxout functions would be further exceptions). If the slope around the origin of \pentan{} is responsible for its good performance, then this could also explain why \cube{} is so bad, since it is very flat close to the origin.  

\paragraph{Influence of hyperparameters} To get some intuition about how hyperparameters affect our different activation functions, we regressed the score of the functions on the test set on all the employed hyperparameters. For example, we estimated:
\begin{align}\label{eq:regression}
  y = \alpha_l\cdot\log(n_l)+\alpha_d\cdot d+\cdots
\end{align}
where $y$ is the score on the test set, $n_l$ is the number of layers in the network, $d$ is the dropout value, etc. The coefficients $\alpha_k$ for each regressor $k$ is what we want to estimate (in particular, their size and their sign). We logarithmized certain variables whose scale was substantially larger than those of others (e.g., number of units, number of filters). For discrete regressors such as the optimizer we used binary dummy variables. We estimated Eq.~\eqref{eq:regression} independently for each activation function and for each mini-experiment. Overall, there was a very diverse pattern of outcomes, preventing us from making too strong conclusions. Still, we observed that while all models performed on average better with fewer hidden layers, particularly \swish{} was robust to more hidden layers (small negative coefficient $\alpha_l$), but also, to a lesser degree, \pentan{}. In the sentence classification tasks, \mysin{} and the maxout functions were particulary robust to an increase of hidden layers. Since \pentan{} is a saturating function and \mysin{} even an oscillating one, we therefore conclude that preserving the gradient (derivative close to one) is not a necessary prerequisite to successful learning in deeper neural networks.



\section{Concluding remarks}\label{sec:conclusion}
%
We have conducted the first large scale comparison of activation functions across several different NLP tasks (and task types) and using different popular neural network types. 
Our main focus was on so-called scalar activation functions, but we also partly included the more costly `many-to-one' 
maxout functions. 

Our findings suggest that the rectifier functions (and the similarly shaped \swish{}) can be top performers for each task, but their performance is unstable and cannot be predicted a priori. One of our major findings is that, in contrast, the saturating \pentan{} function performs much more stably in this respect and can with high probability be expected to perform well across tasks as well as different choices of hyperparameters. This appears to make it the method of choice particularly when hyperparameter optimization is costly. When hyperparameter optimization is cheap, we recommend to consider the activation function as another hyperparameter and choose it, e.g., from the range of functions listed in Table \ref{table:topN} along with maxout. 

Another major advantage of the \pentan{} function is that it may also take the role of a gate (because of its finite range) and thus be employed in more sophisticated neural network units such as LSTMs, where the rectifiers fail completely. In this context, we noticed that replacing \sigmoid{} and \mytanh{} in an LSTM cell with \pentan{} leads to a 2pp increase on a challenging NLP sequence tagging task. Exploring whether this holds across more NLP tasks should be scope for future work. Additionally, our research suggests it is worthwhile to further explore \pentan{}, an arguably marginally known activation function. For instance, other scaling factors than $0.25$ (default value from \citet{Xu:2016}) should be explored. Similarly as for \prelu{}, the scaling factor can also be made part of the optimization problem. 

Finally, we found that except for \swish{} none of the newly discovered activation functions found in \citet{Ramach:2018} made it in our top categories, suggesting that automatic search of activation functions should be made across multiple tasks in the future. 

\section*{Acknowledgments}
We thank Teresa B\"otschen, Nils Reimers and the anonymous reviewers for helpful comments.
This work has been supported by the
German Federal Ministry of Education and Research
(BMBF) under the promotional reference
01UG1816B (CEDIFOR).

\bibliography{emnlp2018-2}
\bibliographystyle{acl_natbib}

\appendix
\clearpage


\section{Supplemental Material}
\begin{figure}[!htb]
\scalebox{0.5}{
\input{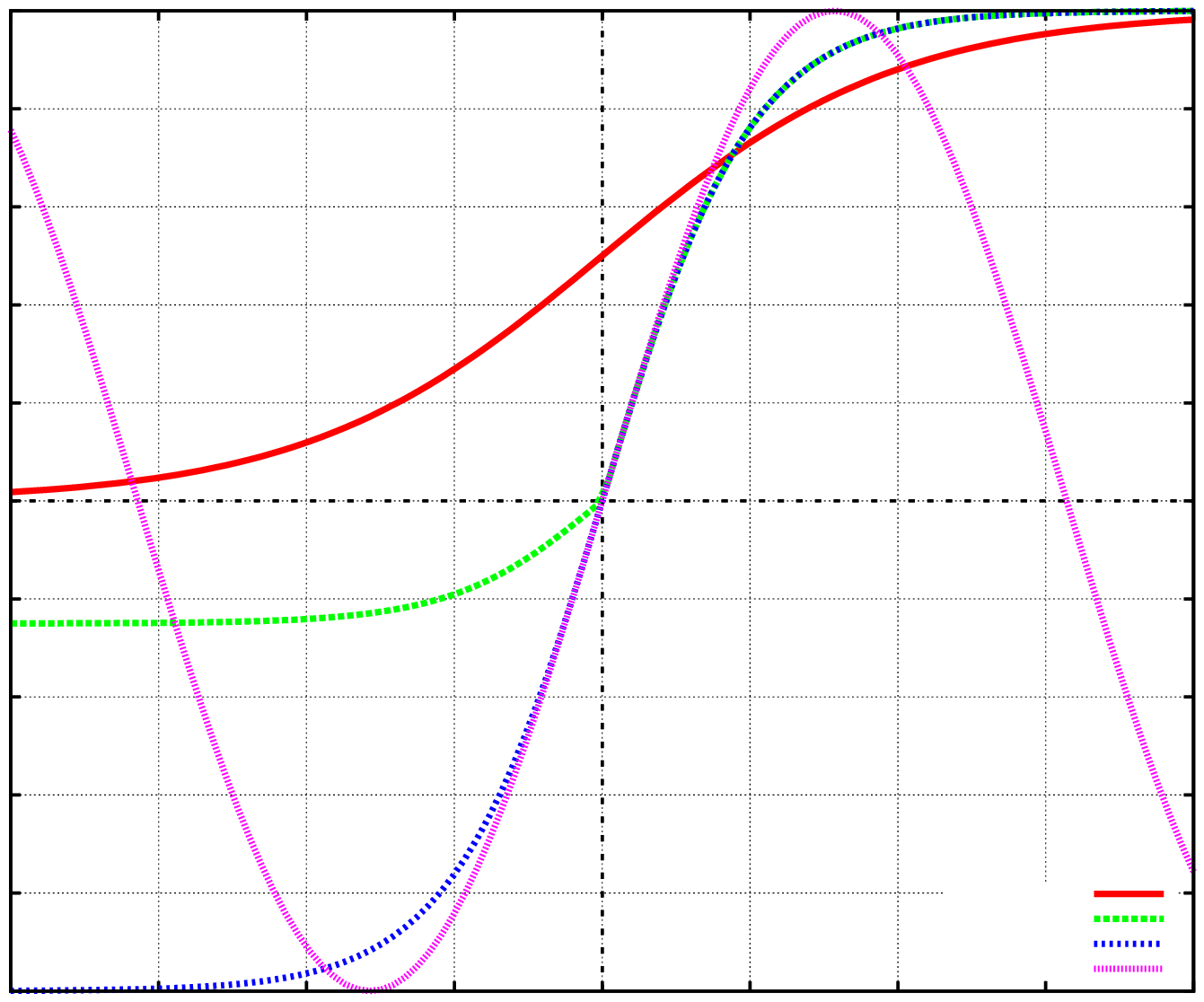}}
\scalebox{0.5}{
\input{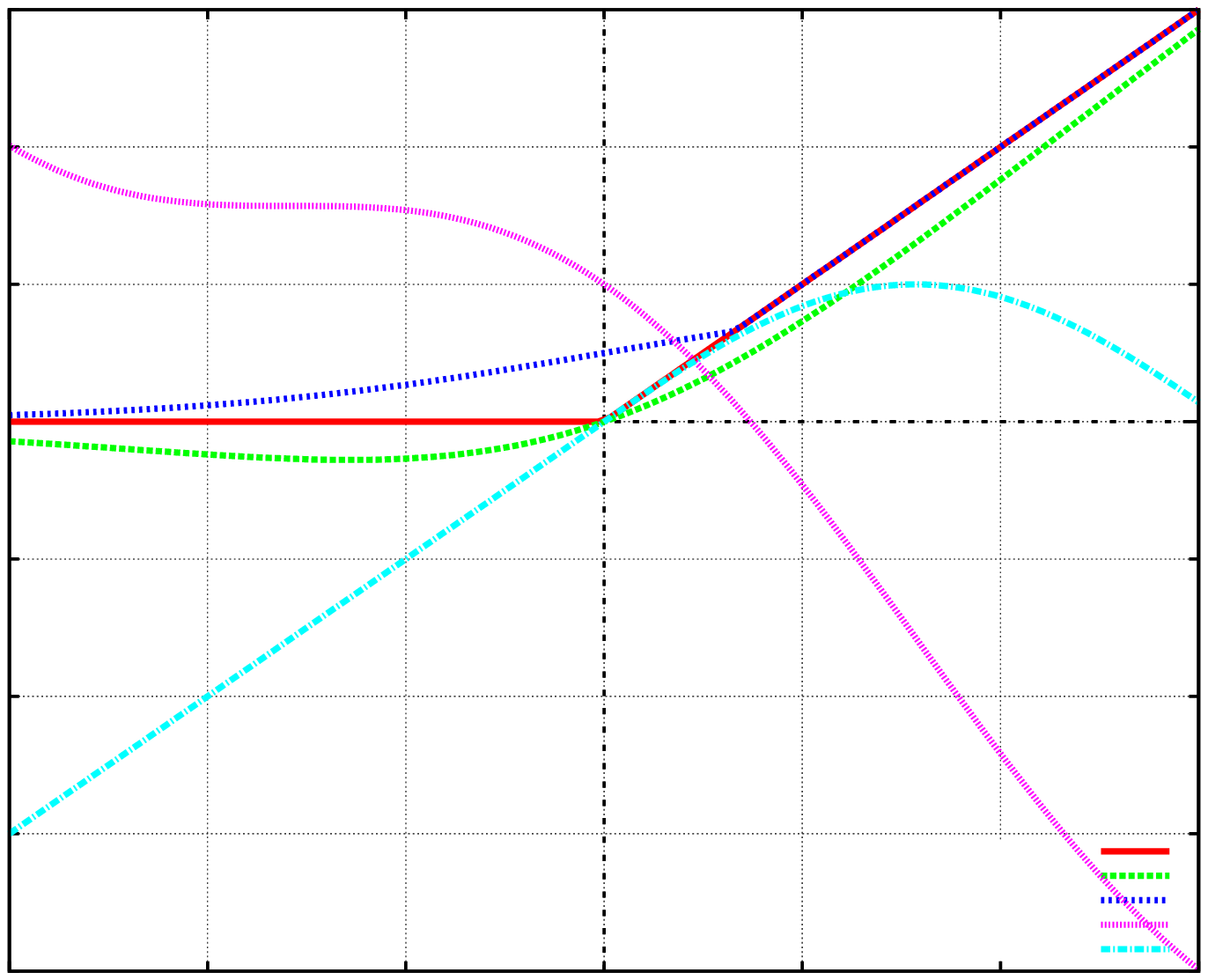}
}
\caption{Several activation functions.}
\label{fig:saturating}
\end{figure}

\begin{figure}[!htb]
\scalebox{0.5}{
\input{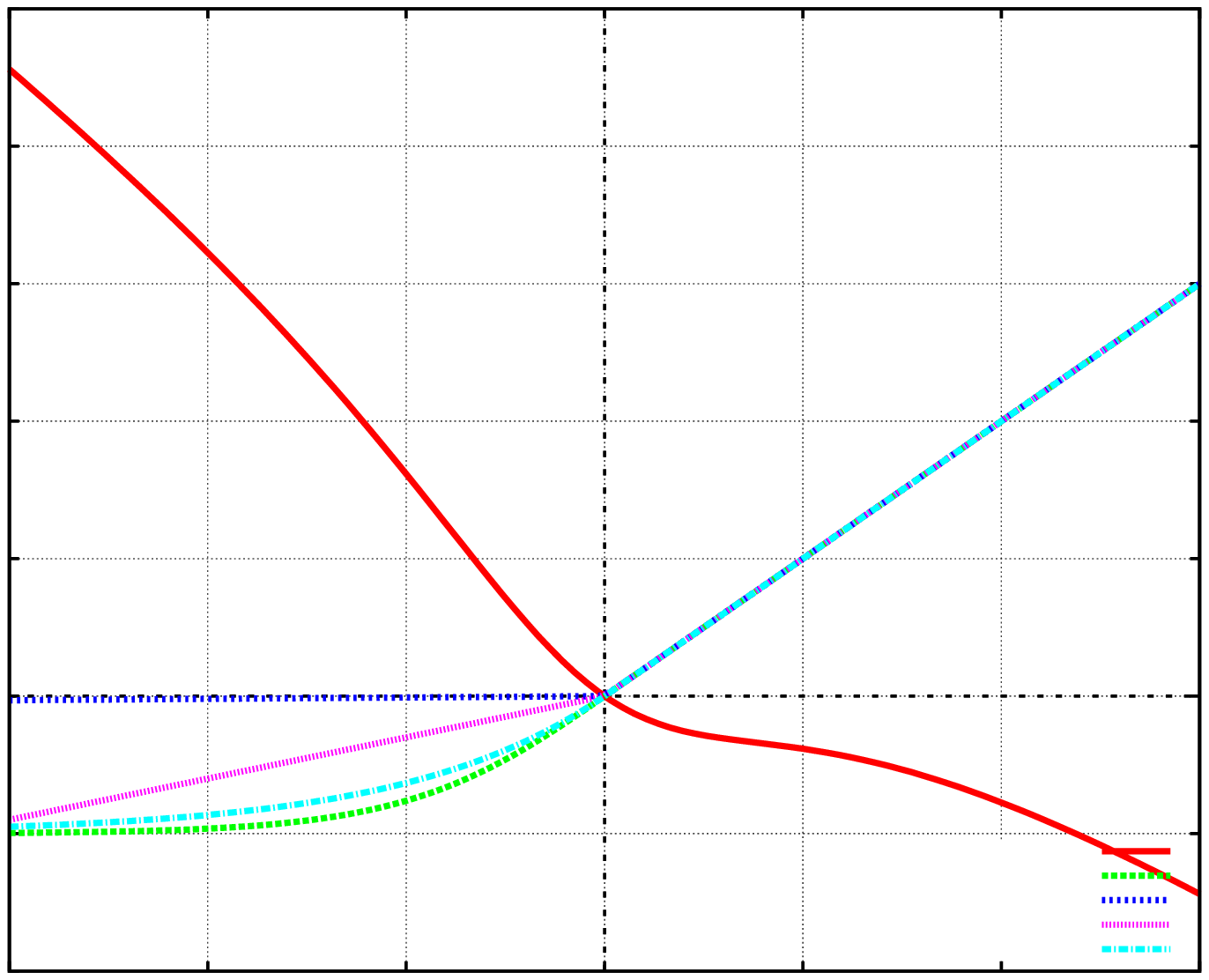}}
\scalebox{0.5}{
\input{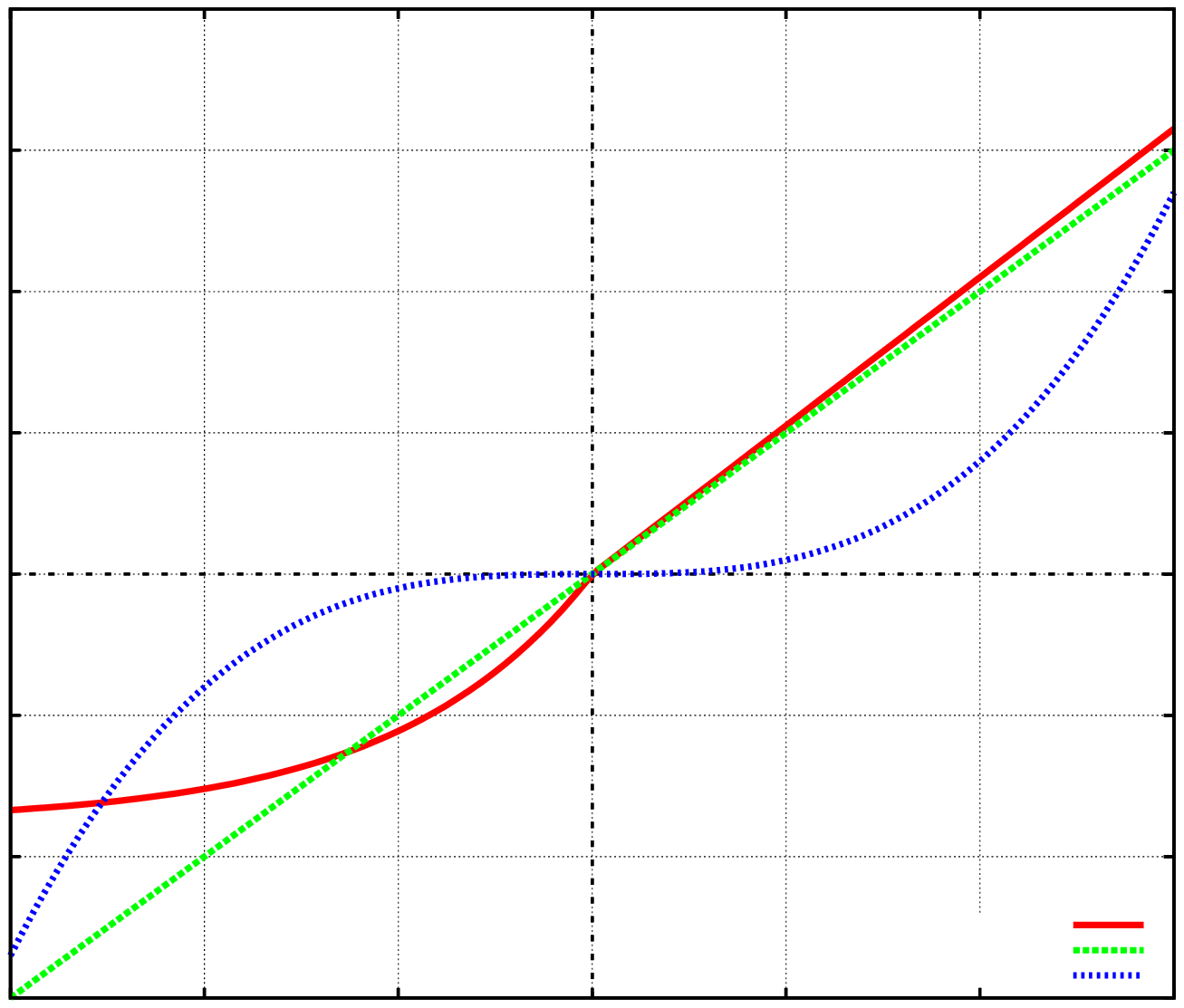}
}
\caption{Several activation functions.}
\label{fig:nonsaturating}
\end{figure}

Figures \ref{fig:saturating} and \ref{fig:nonsaturating} graph the 21 activation functions investigated in this paper.

\end{document}